\documentclass[10pt,twocolumn,letterpaper]{article}

\usepackage[pagenumbers]{cvpr} 

\usepackage{graphicx}
\usepackage{amsmath}
\usepackage{amssymb}
\usepackage{array}
\usepackage{booktabs}
\usepackage{multirow}
\usepackage{threeparttable}
\usepackage{bbding}
\usepackage{makecell}
\usepackage{subcaption}
\usepackage{xcolor}
\usepackage{colortbl}
\usepackage{tcolorbox}
\usepackage{bm}
\usepackage{bbm}
\definecolor{aliceblue}{rgb}{0.94, 0.97, 1.0}
\definecolor{deemph}{gray}{0.6}
\newcommand{\gc}[1]{\textcolor{deemph}{#1}}

\usepackage{pifont}
\newcommand*\colourcheck[1]{%
  \expandafter\newcommand\csname #1check\endcsname{\textcolor{#1}{\ding{52}}}%
}
\colourcheck{blue}
\colourcheck{green}
\colourcheck{cyan}
\colourcheck{teal}
\colourcheck{red}
\newcommand{\xmark}{\textcolor{red}{\ding{55}}}%

\definecolor{cvprblue}{rgb}{0.21,0.49,0.74}
\newlength\savewidth
\usepackage[pagebackref,breaklinks,colorlinks,allcolors=cvprblue]{hyperref}

\def\paperID{*****} 
\def\confName{CVPR}
\def\confYear{2025}

\title{EasyRef: Omni-Generalized Group Image Reference for Diffusion Models via Multimodal LLM}

\newcommand{\authorskip}{\hspace{4mm}}

\author{
 Zhuofan Zong$^{1,2}$ \authorskip Dongzhi Jiang$^{1}$ \authorskip Bingqi Ma$^{2}$ \authorskip Guanglu Song$^{2}$ \\
 Hao Shao$^{1}$ \authorskip Dazhong Shen$^{3}$ \authorskip Yu Liu$^{2}$ \authorskip Hongsheng Li$^{1,3}$ \\\\
 $^1$CUHK MMLab \authorskip $^2$SenseTime Research \authorskip $^3$Shanghai AI Laboratory \\\\
 Project page: \url{https://easyref-gen.github.io/}
}

\begin{document}

\twocolumn[{%
\renewcommand\twocolumn[1][]{#1}%
\maketitle
\begin{center}
    \centering
    \captionsetup{type=figure}
    \includegraphics[width=\linewidth]{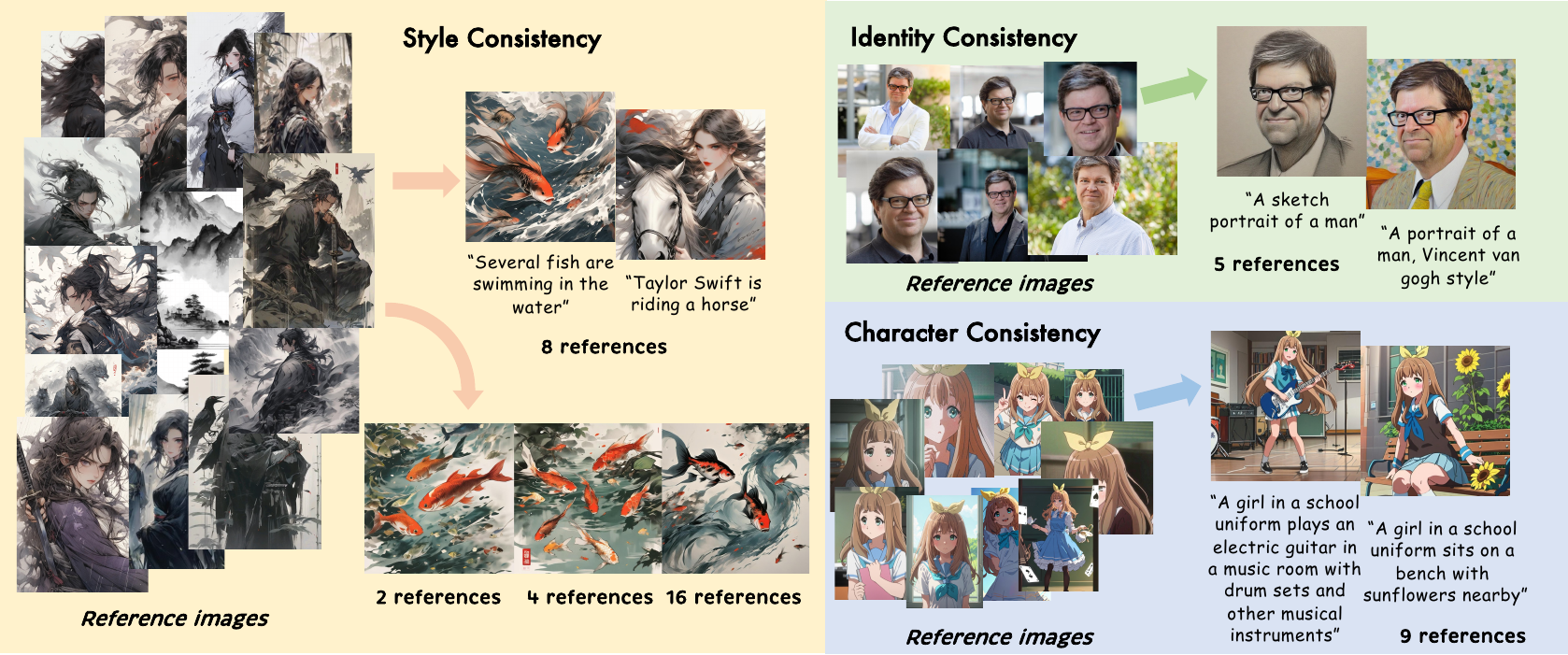}
    \caption{\textbf{EasyRef} is capable of modeling the consistent visual elements of various input reference images with a single generalist multimodal LLM in a zero-shot setting.}
    \label{fig:teaser}
\end{center}%
}]

\renewcommand{\thefootnote}{\fnsymbol{footnote}}

\begin{abstract}
Significant achievements in personalization of diffusion models have been witnessed.
Conventional tuning-free methods mostly encode multiple reference images by averaging their image embeddings as the injection condition, but such an image-independent operation cannot perform interaction among images to capture consistent visual elements within multiple references.
Although the tuning-based Low-Rank Adaptation (LoRA) can effectively extract consistent elements within multiple images through the training process, it necessitates specific finetuning for each distinct image group.
This paper introduces EasyRef, a novel plug-and-play adaptation method that enables diffusion models to be conditioned on multiple reference images and the text prompt.
To effectively exploit consistent visual elements within multiple images, we leverage the multi-image comprehension and instruction-following capabilities of the multimodal large language model (MLLM), prompting it to capture consistent visual elements based on the instruction.
Besides, injecting the MLLM's representations into the diffusion process through adapters can easily generalize to unseen domains, mining the consistent visual elements within unseen data.
To mitigate computational costs and enhance fine-grained detail preservation, we introduce an efficient reference aggregation strategy and a progressive training scheme. 
Finally, we introduce MRBench, a new multi-reference image generation benchmark.
Experimental results demonstrate EasyRef surpasses both tuning-free methods like IP-Adapter and tuning-based methods like LoRA, achieving superior aesthetic quality and robust zero-shot generalization across diverse domains. 
\end{abstract}    
\section{Introduction}
\label{sec:intro}

Significant achievements in diffusion models~\cite{ldm,sdxl,playground3,dalle3,sd3,dalle2,dalle,imagen,comat,raphael,lidit,shen2024rethinking} have been witnessed because of their remarkable abilities to create visually stunning images.
To improve the precision and controllability of diffusion models, researchers have been exploring personalized generation conditioned on a small number of reference images, \textit{i.e.,} the generated images are required to maintain elements of the reference image while incorporating modifications specified by the text prompt.
Such personalized image generation approaches are mainly categorized into tuning-free methods~\cite{ipadapter,instantid,instantstyle,instantstylep,controlnet, cnpp} and tuning-based methods~\cite{lora,dreambooth,textinversion}.

\begin{figure}[tp]
    \centering
    \includegraphics[width=0.9\linewidth]{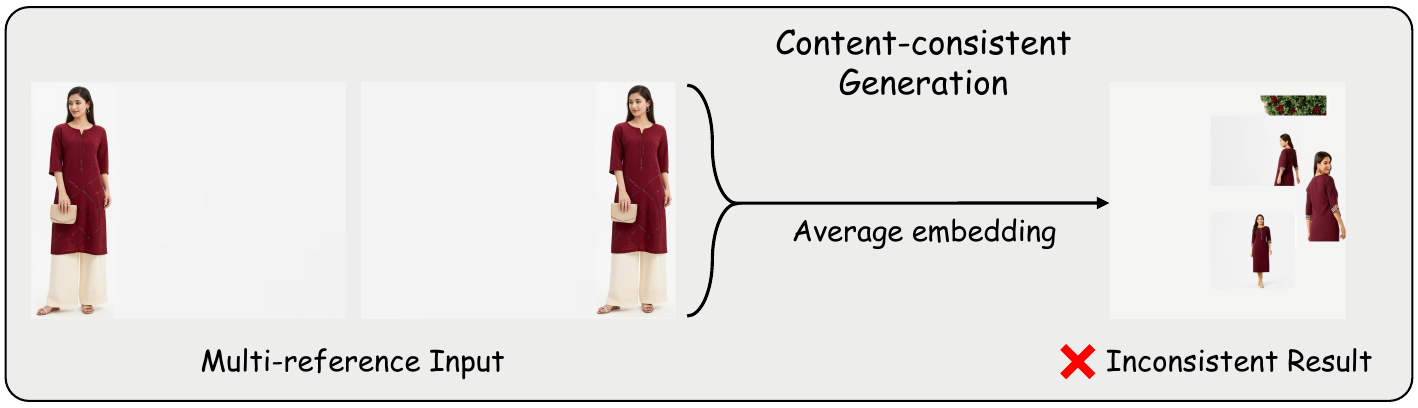}
    \caption{Spatial misalignment issue of the embedding averaging operation. The images with faces are synthetic.}
    \label{fig:average}
    \vspace{-3mm}
\end{figure}

Previous tuning-free approaches typically leverage a pretrained feature extractor to capture certain attributes of the reference image, which are then injected into the frozen diffusion model via trainable adapters.
The seminal IP-Adapter~\cite{ipadapter} proposed this design by integrating CLIP~\cite{clip} embeddings of the reference image with decoupled cross-attention layers.
Follow-up works~\cite{instantid,deadiff,csgo} have developed sophisticated, task-specific feature encoders to encode distinct elements of the reference image (\textit{e.g.}, style, content, character, identity, \textit{etc.}) for personalized image generation.
Despite their promise, these methods have several limitations. 
Firstly, encoders for different references, such as style and character, have specific complex designs and can be optimized through various specialized training tasks. 
Secondly, most methods~\cite{ipadapter,instantstyle,deadiff} are limited to training with a single reference image and fail to fully encode consistent visual representations from multiple references. 
Although Low-Rank Adaptation (LoRA)~\cite{lora} can extract consistent elements within multiple images through the training process, it necessitates specific finetuning for each distinct image group.

This paper introduces EasyRef, a plug-and-play adaption method that empowers diffusion models to condition multiple reference images and text prompts.
Conventional methods mostly encode multiple reference images by averaging their image embeddings as the injection condition, but such an image-independent operation cannot perform interaction among images to capture consistent visual elements within multiple references.
For example, as illustrated in Figure~\ref{fig:average}, the CLIP-based IP-Adapter generates an inconsistent image when the spatial locations of the target subject vary across the reference images.
To effectively exploit consistent visual elements within multiple images, we leverage the multi-image comprehension and instruction-following capabilities of the multimodal large language model (MLLM)~\cite{llava,llavanext,blip,blip2,instructblip,mova,visualcot,internvl,internvl15}, prompting it to capture consistent visual elements based on the instruction.
Besides, injecting the MLLM's representations into the diffusion process through adapters can easily generalize to unseen domains, mining the consistent visual elements within unseen data.
EasyRef also inherits the MLLM's ability to process arbitrary number of reference images with arbitrary aspect ratios.
We present the key differences among our method, LoRA, and IP-Adapter in Table~\ref{tab:intro}.
To mitigate the computational demands imposed by the long context of multi-image inputs, we propose querying the MLLM with learned token embeddings and aggregating reference representations within the deepest layer of the MLLM architecture.
Additionally, to address the limitations of MLLMs in capturing fine-grained visual details, we employ a progressive training strategy to enhance the MLLM's capacity for fine-grained detail and identity preservation. 
Unlike previous methods that rely on sophisticated feature encoders~\cite{clip,dinov2,deadiff} and even additional face encoders~\cite{arcface}, we find the single MLLM in EasyRef is capable of extracting various reference representations, such as style and character, along with text features from an arbitrary set of reference images and the text prompt, exhibiting strong generalization ability.
Finally, we introduce a multi-reference generation benchmark (MRBench) for multi-reference image generation to evaluate our work and guide future research. Compared to IP-Adapter and the prevalent fine-tuning approaches using LoRA, our proposed generalist EasyRef model consistently achieves superior aesthetic quality across diverse domains and demonstrates robust zero-shot generalization.

\begin{table}[tp]
\centering
\resizebox{0.99\linewidth}{!}{
\begin{tabular}{@{}lccc@{}}
    \toprule
    Method   & Consistency mining & Zero-shot generalization & Multi-reference input \\ 
    \midrule
    LoRA~\cite{lora} & \cyancheck & \xmark & \xmark \\ 
    IP-Adapter~\cite{ipadapter} & \xmark & \cyancheck & \cyancheck \\
    EasyRef & \cyancheck & \cyancheck & \cyancheck \\
    \bottomrule
\end{tabular}
    }
\caption{Comparison among LoRA, IP-Adapter, and EasyRef.}
\label{tab:intro}
\vspace{-2mm}
\end{table}

In summary, our contributions are threefold: (1) We introduce EasyRef, the first plug-and-play adaptation technique enabling diffusion models to be jointly conditioned on multiple reference images and text prompts. (2) We propose an efficient reference aggregation strategy and a progressive training scheme to mitigate computational costs and enhance the MLLM's fine-grained perceptual abilities. (3) We introduce a novel MRBench for evaluating diffusion models in multi-reference image generation scenarios.
\section{Related Work}
\label{sec:related}


\subsection{Image Personalization}
Image personalization approaches can be categorized into tuning-free methods~\cite{ipadapter,instantid,instantstyle,instantstylep,controlnet, cnpp, comat,imagineyourself,deadiff,photomaker} and tuning-based methods~\cite{lora,dreambooth,textinversion}.
Tuning-free approaches typically extract visual representations, such as style and character, from the reference image and inject these into the diffusion model.
IP-Adapter~\cite{ipadapter} enhances image prompting capabilities through a decoupled cross-attention mechanism.
Building upon IP-Adapter, InstantStyle~\cite{instantstyle,instantstylep} injects CLIP~\cite{clip} style embeddings into style-specific blocks.
Both IP-Adapter-Face~\cite{ipadapter} and InstantID~\cite{instantid} employ additional face encoders~\cite{arcface} to improve identity preservation.
A limitation of tuning-free methods is that they are trained with single-reference input, failing to fully exploit the consistent elements within multiple reference images.
Tuning-based approaches, such as LoRA~\cite{lora}, finetuned the diffusion model using a limited set of images.
Although tuning-based methods are capable of multi-image references, a key limitation is they necessitate specific finetuning for each distinct image group.
In this work, we extend tuning-free methods to accommodate multiple reference images and the text prompt like tuning-based methods while maintaining robust generalization capabilities.

\subsection{Multimodal Large Language Models}
Multimodal large language models (MLLMs)~\cite{llava,llavanext,blip,blip2,instructblip,mova,visualcot,internvl,internvl15} have demonstrated remarkable success in addressing open-world tasks.
Pioneering works like LLaVA~\cite{llava} and BLIP-2~\cite{blip2} consistently projected the vision representation from a pretrained CLIP vision encoder into the LLM for multimodal comprehension.
Qwen-VL~\cite{qwenvl} collected massive multimodal tuning data and adopted elaborate training strategy for better optimization.
The mixture-of-vision-experts designs, such as SPHINX~\cite{sphinx}, MoF~\cite{mof}, and MoVA~\cite{mova}, were explored to enhance the visual capabilities of MLLMs.
Furthermore, models like LLaVA-NeXT~\cite{llavanext} and Qwen2-VL~\cite{Qwen2VL} sought to enable the processing of images with arbitrary resolutions.
LI-DiT~\cite{lidit} investigated how to effectively unleash the MLLM's prompt encoding capabilities for diffusion models.
In this paper, we are the first to leverage the multi-image comprehension and instruction-following capabilities of the MLLM to jointly encode representations of multiple reference images and the text prompt.
\section{EasyRef}

\begin{figure*}[tp]
    \centering
    \includegraphics[width=0.9\textwidth]{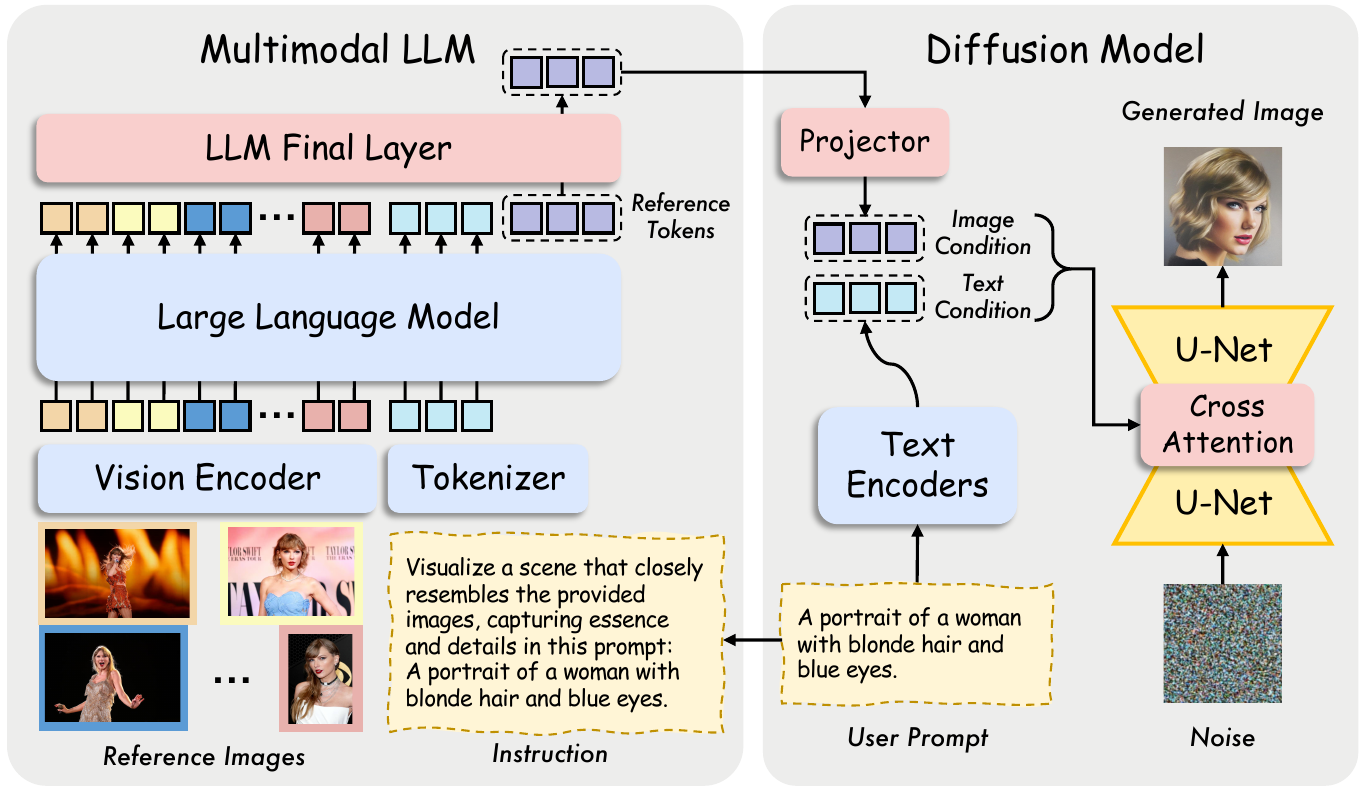}
    \caption{\textbf{Overview of EasyRef with SDXL.} EasyRef extracts consistent visual eliments from multiple reference images and the text prompt via a MLLM, injecting the condition representations into the diffusion model through cross-attention layers. We only plot 1 cross-attention layer for simplicity.}
    \label{fig:framework}
    \vspace{-5mm}
\end{figure*}

\subsection{Preliminary}
Denoising Diffusion Probabilistic Models~\cite{ddpm} (DDPMs) are trained by maximizing the log-likelihood of the training data, given a data distribution  \(q(\mathbf{x}_0)\). The training process involves a forward diffusion process that gradually adds Gaussian noise to the data over \(T\) timesteps:

\begin{equation}
    q(\mathbf{x}_{1:T} | \mathbf{x}_0) = \prod_{t=1}^T q(\mathbf{x}_t | \mathbf{x}_{t-1}),
\end{equation}
\begin{equation}
    q(\mathbf{x}_t | \mathbf{x}_{t-1}) = \mathcal{N}(\mathbf{x}_t; \sqrt{{\alpha}_t} \mathbf{x}_{t-1}, (1 - {\alpha}_t) \mathbf{I}).
\end{equation}
Here, \(\mathbf{x}_t\) represents the noisy data at timestep \(t\) and \({\alpha}_t\) is a schedule parameter controlling the noise level at each timestep. 
The core of DDPM training lies in learning a parameterized model \(p_\theta\) to approximate the reverse diffusion process:

\begin{equation}
     p_\theta(\mathbf{x}_{0:T}) = p(\mathbf{x}_T) \prod_{t=1}^T p_\theta(\mathbf{x}_{t-1} | \mathbf{x}_t),
\end{equation}
\begin{equation}
    p_\theta(\mathbf{x}_{t-1} | \mathbf{x}_t) = \mathcal{N}(\mathbf{x}_{t-1}; \boldsymbol{\mu}_\theta(\mathbf{x}_t, t), \boldsymbol{\sigma}^2_t \mathbf{I}).
\end{equation}
This model learns to progressively remove noise from a given noisy sample \(\mathbf{x}_t\), recovering the original data \(\mathbf{x}_0\).  


\subsection{Methodology}
As illustrated in Figure~\ref{fig:framework}, EasyRef comprises four key components: (1) a pretrained diffusion model for conditional image generation, (2) a pretrained multimodal large language model (MLLM) for encoding a set of reference images and the text prompt, (3) a condition projector that maps the representations from the MLLM into the latent space of diffusion model, and (4) trainable adapters for integrating image conditioning embedding into the diffusion process.

\vspace{1mm}
\noindent\textbf{Reference representation encoding.}
Existing mainstream approaches~\cite{ipadapter,deadiff,instantstyle,instantid} mostly average the CLIP image embeddings of all reference images as the reference condition.
This image-independent operation cannot effectively capture consistent visual elements among reference images.
It also fails to jointly encode text and causes the spatial misalignment issue as presented in Figure~\ref{fig:average}.
To alleviate this issue, we propose to leverage the multi-image comprehension and instruction-following capabilities of the MLLM to encode multi-reference inputs and the text prompt based on the instruction.
We adopt the state-of-the-art Qwen2-VL-2B as our MLLM in this work.
The MLLM consists of a $l$-layer large language model (LLM) and a vision encoder capable of handling images with arbitrary resolutions.
The input image is initially converted into visual tokens
with the vision encoder.
Then we employ an instruction and integrate all images into the instruction, which explicitly encourages the MLLM to focus on the crucial and common contents within the reference images.
These multimodal input tokens are subsequently processed by the LLM.

\vspace{1mm}
\noindent\textbf{Efficient reference aggregation.} 
Increasing the number of reference images inevitably raises the number of visual tokens in the LLM. 
This extended context length substantially elevates the computational cost for the diffusion model. 
We propose to encapsulate the reference representations into $N$ learnable reference tokens $\mathbf{F}_{\textrm{ref}} \in \mathbb{R}^{N \times D}$ in the LLM to achieve efficient inference.
However, all parameters of LLM must be trained to interpret these newly added tokens. 
To enhance training efficiency, we append $\mathbf{F}_{\textrm{ref}}$ to the context sequence $\mathbf{F}_{l-1}$ at the final layer of LLM, keeping all previous LLM layers frozen during pretraining:
\begin{equation}
    {\mathbf{F}}_{l}' = \text{Concat}(\mathbf{F}_{l-1}, \mathbf{F}_{\textrm{ref}})
\end{equation}
We then adopt bi-directional self-attention to facilitate the propagation of representations across the reference images in the final layer, followed by a multi-layer perception network (MLP):
\begin{equation}
    \mathbf{F}_{l}'' = \text{MLP}(\text{Bi-Attention}({\mathbf{F}}_{l}')),
\end{equation}
where we omit the residual addition in attention and MLP layers for simplicity.
Next, we split \( \mathbf{F}_{l}'' \) into the updated representations \( \mathbf{F}_{l} \) and the encapsulated reference tokens \( \mathbf{F}_{\text{ref}}' \):
\begin{equation}
    \mathbf{F}_{l},\mathbf{F}_{\textrm{ref}}' = \text{Split}(\mathbf{F}_{l}'').
\end{equation}
Finally, we project \( \mathbf{F}_{\text{ref}}' \) through a trainable MLP condition projector to obtain the final conditioning vector \( \mathbf{c}_{i} \):
\begin{equation}
    \mathbf{c}_{i} = \text{MLP}(\mathbf{F}_{\textrm{ref}}'),
\end{equation}

\vspace{1mm}
\noindent\textbf{Reference representation injection.} 
The text conditions are injected into the pretrained diffusion model through cross-attention layers.
Following IP-Adapter, we introduce a new cross-attention layer into each cross-attention layer of the U-Net.
Given the latent features $\mathbf{X}$, text conditions $\mathbf{c}_t$, and image conditions $\mathbf{c}_i$, the injected features $\hat{\mathbf{X}}$ are computed by the cross-attention layer as follows:
\begin{equation}
    \hat{\mathbf{X}} = \text{Softmax}\left(\frac{\mathbf{Q}\mathbf{K}^{T}}{\sqrt{d}}\right)\mathbf{V} + 
    \text{Softmax}\left(\frac{\mathbf{Q}\hat{\mathbf{K}}^{T}}{\sqrt{d}}\right)\hat{\mathbf{V}},
\end{equation}
where $\hat{\mathbf{K}}=\mathbf{c}_i\hat{\mathbf{W}}_k$ and $\hat{\mathbf{V}}=\mathbf{c}_i\hat{\mathbf{W}}_v$.
Both $\hat{\mathbf{W}}_k$ and $\hat{\mathbf{W}}_v$ are newly added trainable parameters.

\subsection{Progressive Training Scheme}

\vspace{1mm}
\noindent\textbf{Alignment pretraining.}
To facilitate the adaption of MLLM's visual signals to the diffusion model, we construct a large-scale dataset containing 13M high-quality image-text pairs, including LAION-5B~\cite{laion} and other internal datasets for the alignment pretraining.
During the pretraining phase, we only optimize the final layer and reference tokens of the MLLM along with the newly added adapters and condition projector while preserving the capabilities of the initial MLLM and diffusion model.
The shorter side of the input image is resized to 1024 and we further center crop 1024 $\times$ 1024 pixels of the image. 

\vspace{1mm}
\noindent\textbf{Single-reference finetuning.}
Following alignment pretraining, the MLLM is trainable and subjected to single-reference fine-tuning.
Specifically, we subsequently unfreeze the vision encoder and all layers of the MLLM to enhance its capacity for fine-grained visual perception at the second stage. 
We additionally incorporate trainable Low-Rank Adaption (LoRA) layers to attention layers of the frozen U-Net.
Building upon the aforementioned pre-training dataset, we augment the training data with 4M real-world human images from LAION-5B, utilizing cropped face regions as conditioning inputs. 
The training resolution setting keeps consistent with the first stage.

\vspace{1mm}
\noindent\textbf{Multi-reference finetuning.}
The third stage enables the MLLM to accurately comprehend the common elements across multiple image references and generate high-quality, consistent images. 
Training is performed on a curated dataset comprising image groups, where each group contains multiple images of the same topic (\textit{e.g.}, Donald Trump, a Tesla Model 3, \textit{etc}.) with varying aspect ratios. 
During training, one image from each group is randomly selected as the optimization target, while the remaining ones serve as the conditioning inputs. 
Data augmentation, including random shuffling and truncation, is applied to the conditioning images.
We keep the original aspect ratio for each target image.

\vspace{1mm}
\noindent\textbf{Training supervision.}
We use the same training objective as the original stable diffusion model:
\begin{equation}
\mathcal{L} = \mathbb{E}_{\mathbf{x}_0, \epsilon, \mathbf{c}_t, \mathbf{c}_i, t} \left\| \epsilon - \epsilon_\theta(\mathbf{x}_t, \mathbf{c}_t, \mathbf{c}_i, t) \right\|^2,  
\end{equation}
where $\mathbf{c}_t$ and $\mathbf{c}_i$ denote the text condition and image condition, respectively.

\begin{figure}[tp]
    \centering
    \includegraphics[width=0.75\linewidth]{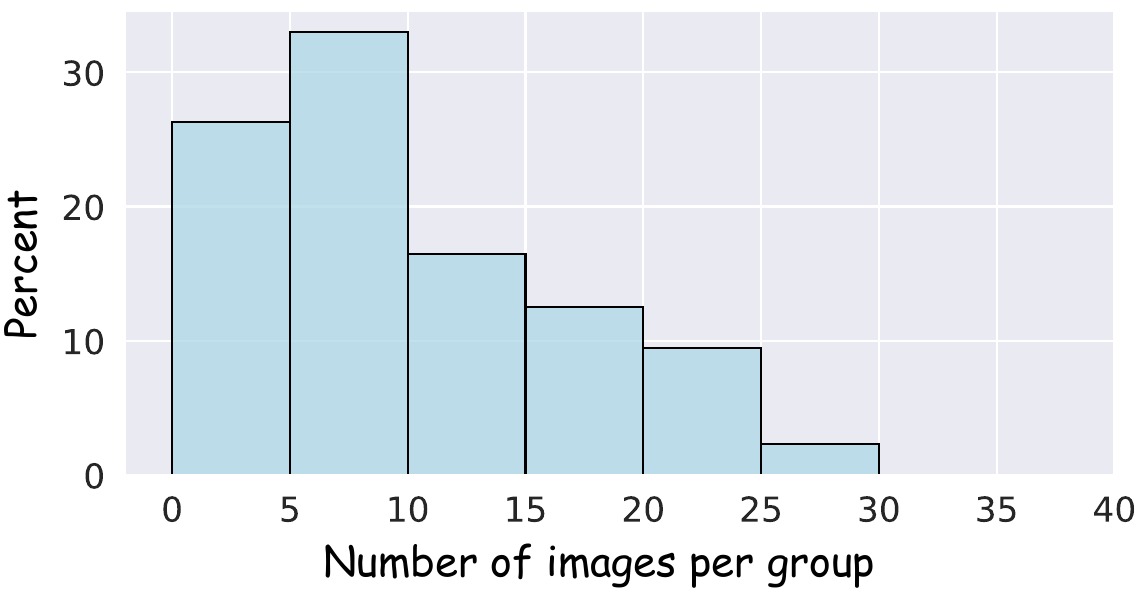}
    \caption{Distribution of our curated dataset.}
    \label{fig:distribution}
    \vspace{-3mm}
\end{figure}

\section{Multi-Reference Generation Benchmark}

\begin{figure*}[tp]
    \centering
    \includegraphics[width=0.999\textwidth]{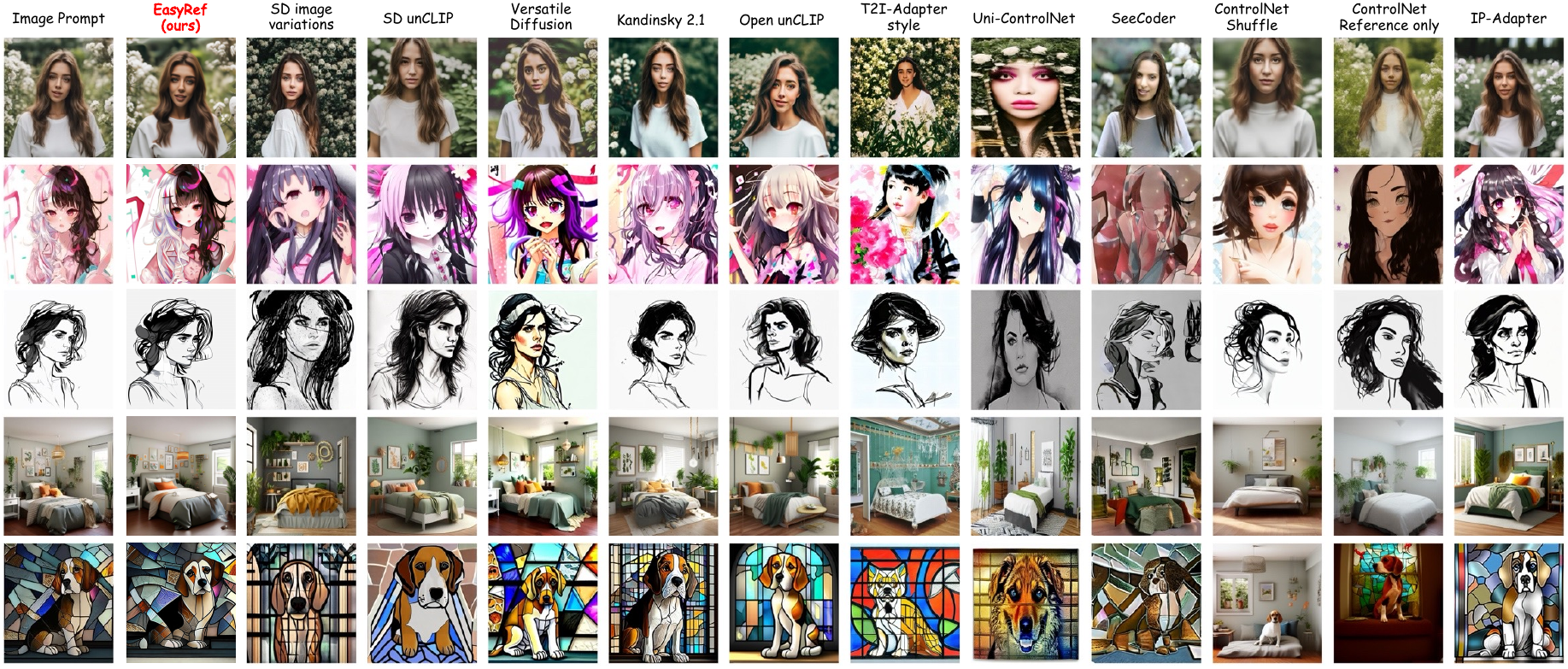}
    \caption{Comparisons of EasyRef with other counterparts in various single-image reference scenarios. The same image prompts as in~\cite{ipadapter} are used for clear comparisons.}
    \label{fig:single_ref}
    \vspace{-3mm}
\end{figure*}

\vspace{1mm}
\noindent\textbf{Dataset construction.}
We constructed a tag list including celebrities, characters, styles, and common objects, then collected images from diverse sources based on the list.
Images sharing the same tag are set into the same group.
To generate aligned text captions of the images, synthetic captions generated by Qwen2-VL-7B using the instruction ``\textit{Give a brief, concise and precise caption for this image.}'' were adopted for each image-text pair. 
The resulting dataset comprises 8,912,439 images organized into 1,075,378 groups.
We set the maximum and minimum group sizes as 28 and 2, respectively, ensuring a balanced group size distribution within the dataset.
Figure~\ref{fig:distribution} illustrates the group distribution of our curated dataset.

\vspace{1mm}
\noindent\textbf{Data filtering.}
We also employ a series of efforts for data cleaning.
First, low-resolution images and those with low aesthetic scores were excluded.
Then we filter out image-text pairs with low CLIP image-text similarity scores using CLIP ViT-L/14.
To ensure the consistency within each group, we compute the average CLIP-I and DINO-I scores for each image relative to the other images within its group and filter images with low scores.
Finally, recognizing that conventional filtering methods may not adequately identify specific patterns, such as image collages or high-quality images containing dense text, we manually annotate a subset of our collected samples and train a CLIP-based binary classifier to effectively score and filter these instances. 

\vspace{1mm}
\noindent\textbf{Benchmark splits.}
The collected image-text pairs are divided into the training dataset, the held-in evaluation set, and the held-out evaluation set.
We first sample 60 image groups to construct the held-out evaluation set to evaluate the model performance on unseen data.
The number of images in each set varies.
For each group, each image can be chosen as the target image and others are regarded as the reference images.
There are a total of 487 images in the held-out evaluation set.
To compare our method with multi-reference generation approaches that require finetuning (\textit{e.g.}, LoRA), we randomly selected 300 groups from the remaining 1,075,318 groups to form a test set of 2063 samples.
Unlike the held-out set, only a randomly selected image serves as the target image in each group.
All reference images of the held-in split and other 1,075,018 groups construct the training set.
To improve the aesthetic quality of generated images, only images with aesthetic scores higher than 5.5 can be used as the target images during training.
There are 1,726,763 valid target images in the training set, with an average of 1.6 images per group.

\vspace{1mm}
\noindent\textbf{Evaluation protocol.}
When evaluating the held-in and held-out sets, we use the reference images and the target prompt to generate two images for each group.
Then we employ conventional metrics, including CLIP-I, CLIP-T, and DINO-I, to measure the alignment between generated images and the corresponding target images or prompts.
We mainly consider CLIP-I and DINO-I for the image-image alignment, which computes the similarities of image embeddings from CLIP ViT-L/14 and DINOv2-Small~\cite{dinov2}.
For the image-text alignment, we adopt the CLIPScore~\cite{clipscore}.

\section{Experiments}

\begin{table}[!t]
\centering
\resizebox{0.99\linewidth}{!}{
\begin{tabular}{@{}lccc@{}}
    \toprule
    Method   & CLIP-I $\uparrow$ & CLIP-T $\uparrow$ & DINO-I $\uparrow$ \\  
    \midrule
    \multicolumn{4}{c}{\it Training from scratch} \\
    \midrule
    Open unCLIP~\cite{dalle2}  & 0.858 & 0.608 & - \\
    Kandinsky-2-1~\cite{kandinsky} & 0.855 & 0.599 & - \\
    Versatile Diffusion~\cite{vd} & 0.830 & 0.587 & - \\
    \midrule
    \multicolumn{4}{c}{\it Finetuning} \\
    \midrule    
    SD Image Variations & 0.760 & 0.548 & - \\
    SD unCLIP & 0.810 & 0.584 & - \\
    \midrule
    \multicolumn{4}{c}{\it Adapters} \\
    \midrule    
    Uni-ControlNet~\cite{unicontrol}~(Global Control) & 0.736 & 0.506 & - \\
    T2I-Adapter~\cite{t2iadapter}~(Style) & 0.648  & 0.485 & - \\
    ControlNet Shuffle~\cite{controlnet} & 0.616  & 0.421 & - \\ 
    IP-Adapter$^*$~\cite{ipadapter} & 0.828 & 0.588 & - \\
    IP-Adapter-SDXL$^*$~\cite{ipadapter} & 0.836 & 0.617 & 0.650 \\
    \cellcolor{aliceblue}EasyRef & \cellcolor{aliceblue}\textbf{0.876} & \cellcolor{aliceblue}\textbf{0.621} & \cellcolor{aliceblue}\textbf{0.873} \\
    \bottomrule
\end{tabular}
}
\caption{Evaluation for generation conditioned by COCO validation images. Methods with * use CLIP embeddings and tend to achieve higher scores of CLIP-based metrics due to its preference.}
\vspace{-3mm}
\label{tab:coco}
\end{table}

\subsection{Implementation Details}
\vspace{1mm}
\noindent\textbf{Training.}
We build our EasyRef framework with the established Stable Diffusion XL~\cite{sdxl} model, utilizing the state-of-the-art Qwen2-VL-2B~\cite{Qwen2VL} as the MLLM. 
The resolution of an input image with arbitrary aspect ratio processed by the MLLM can not exceed 336 $\times$ 336.
We introduce 64 reference tokens in the MLLM.
Similar to IP-Adapter, we employ a drop probability of 0.05 for both text and image prompts independently, and a joint drop probability of 0.05 for simultaneous removal of both modalities.
We simply treat a square black image as the empty image condition if the image condition is dropped.
For the implementation of LoRA comparison, we fine-tuned the model using the reference images and employed a LoRA rank of 32.
We present more results in the Appendix.

\begin{table}[!t]
\centering
\resizebox{0.8\linewidth}{!}{
\begin{tabular}{@{}lccc@{}}
    \toprule
    Method   & CLIP-I $\uparrow$ & CLIP-T $\uparrow$ & DINO-I $\uparrow$ \\  
    \midrule
    \multicolumn{4}{c}{\it Held-in split} \\
    \midrule
    LoRA~\cite{lora} & 0.831 & 0.715 & 0.654 \\ 
    \gc{IP-Adapter-SDXL~\cite{ipadapter}} & \gc{0.768} & \gc{0.632} & \gc{0.527} \\ 
    \cellcolor{aliceblue}EasyRef & \cellcolor{aliceblue}\textbf{0.843} & \cellcolor{aliceblue}\textbf{0.726} & \cellcolor{aliceblue}\textbf{0.672} \\
    \midrule
    \multicolumn{4}{c}{\it Held-out split} \\
    \midrule    
    \gc{LoRA~\cite{lora}} & \gc{failed} & \gc{failed} & \gc{failed} \\ 
    IP-Adapter-SDXL~\cite{ipadapter} & 0.795 & 0.645 & 0.579 \\ 
    \cellcolor{aliceblue}EasyRef & \cellcolor{aliceblue}\textbf{0.833} & \cellcolor{aliceblue}\textbf{0.709} & \cellcolor{aliceblue}\textbf{0.614} \\
    \bottomrule     
\end{tabular}
    }
    
\caption{Evaluation for multi-reference image generation on MRBench. ``failed'' means LoRA fails to generalize to the unseen held-out split in a zero-shot setting.}
\label{tab:mrbench}
\vspace{-2mm}
\end{table}
\begin{figure}[tp]
    \centering
    \includegraphics[width=0.9\linewidth]{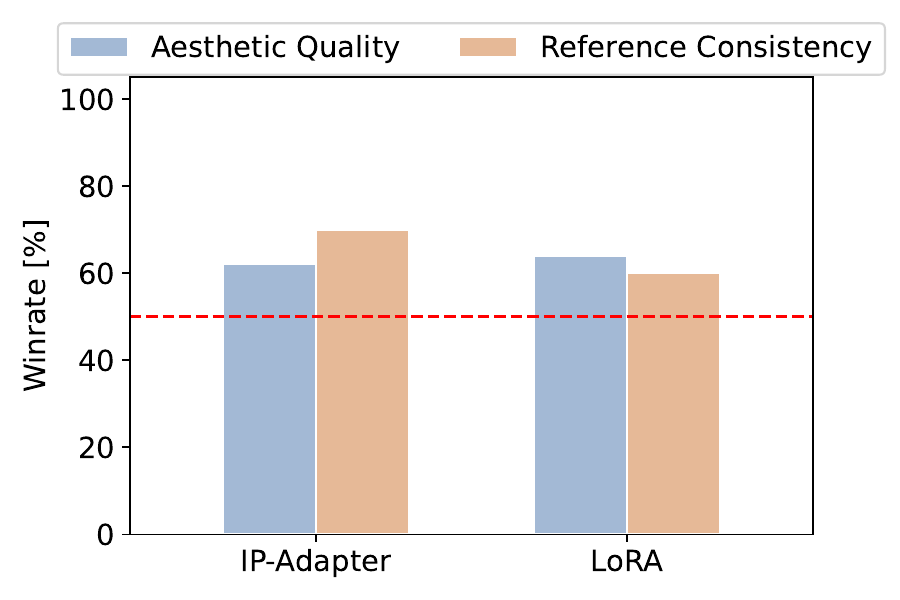}
    \caption{Comparisons of human preference evaluation on our MRBench. EasyRef can surpass other methods across the aesthetic quality and reference alignment.}
    \label{fig:human_eval}
    \vspace{-5mm}
\end{figure}

\vspace{1mm}
\noindent\textbf{Evaluation.}
During inference, we leverage a DDIM~\cite{ddim} sampler with 30 steps and a guidance scale~\cite{cfg} of 7.5. 
As the original IP-Adapter does not support multi-image references, we employed the average of the CLIP embeddings as the image conditioning input.

\begin{figure*}[tp]
    \centering
    \includegraphics[width=1.0\textwidth]{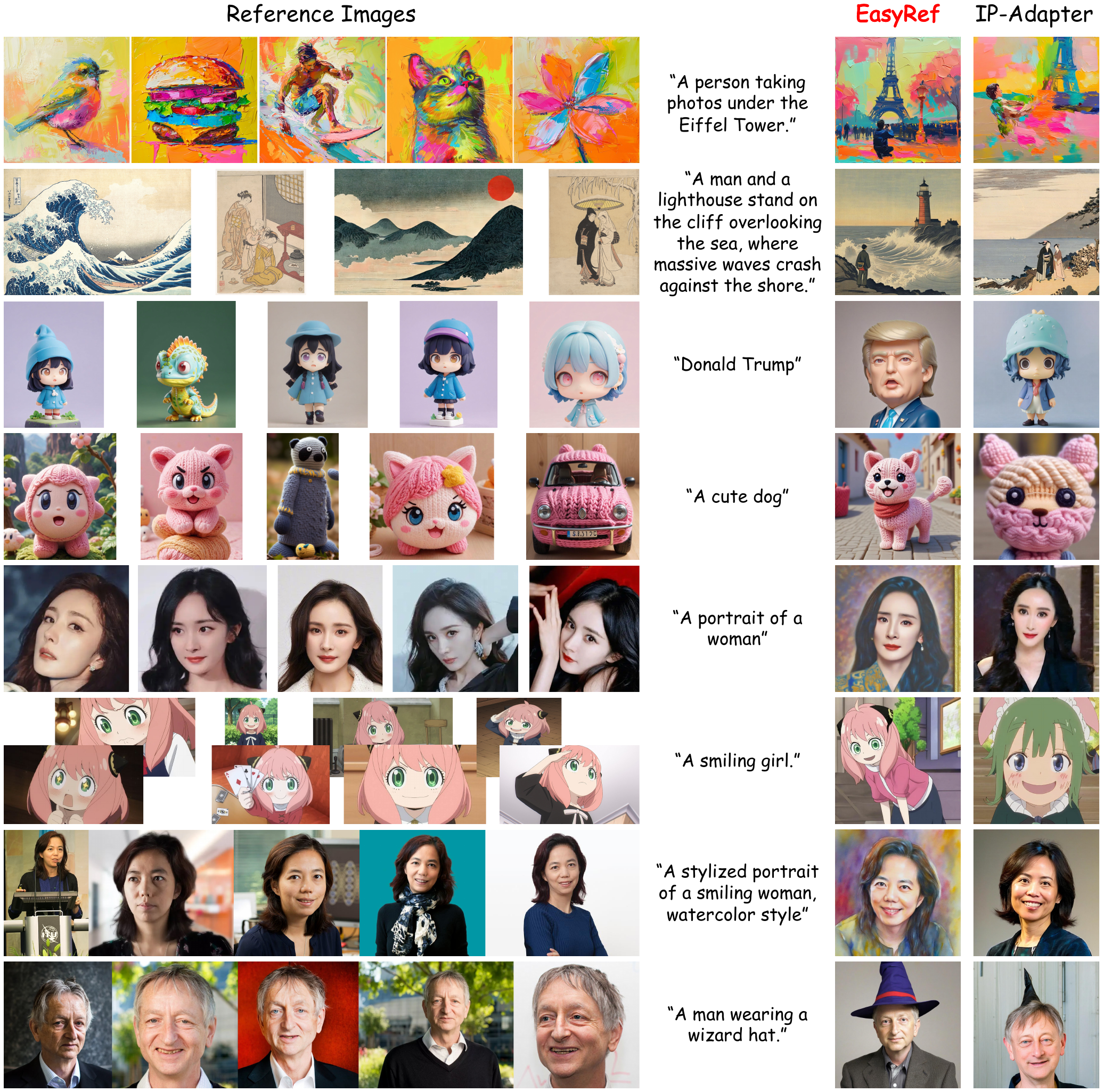}
    \caption{Visualization of generated samples with various multi-reference inputs. These reference contents encompass style, identity, and character, and are encoded by a single generalist MLLM in EasyRef.}
    \label{fig:multi_ref}
    \vspace{-1mm}
\end{figure*}
\begin{figure*}[tp]
    \centering
    \includegraphics[width=0.95\textwidth]{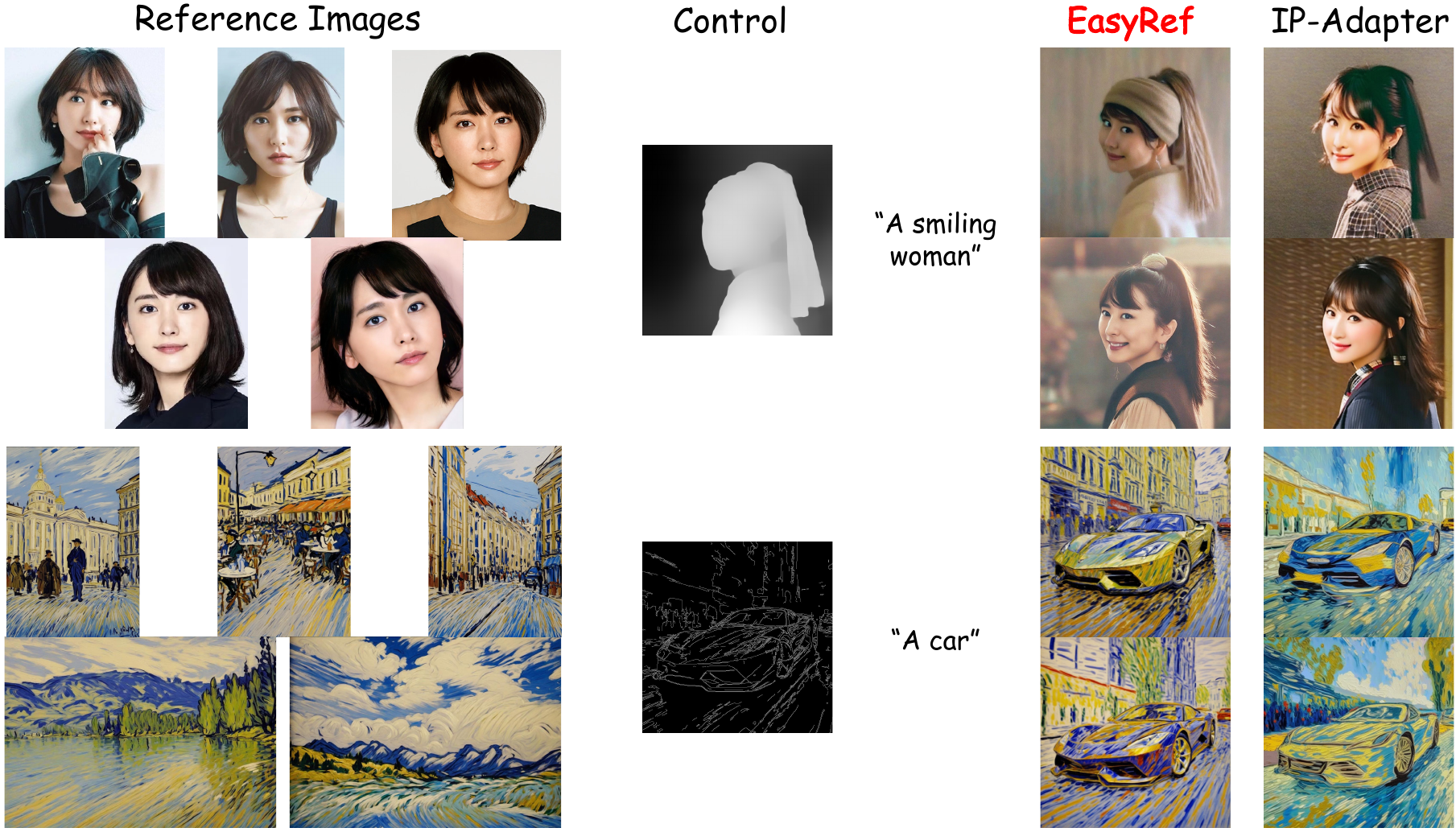}
    \caption{Comparison between EasyRef and IP-Adapter-SDXL with additional structure controls.}
    \label{fig:controlnet}
\end{figure*}

\subsection{Quantitative and Qualitative Results}
\vspace{1mm}
\noindent\textbf{Single-image reference.}
We quantitatively compare our method with other counterparts in single-reference scenarios using the COCO 2017 validation dataset~\cite{mscoco}, which comprises 5000 image-text pairs. 
We use the checkpoint trained by single-reference finetuning.
As shown in Table~\ref{tab:coco}, EasyRef consistently outperforms other methods in both CLIP-T and DINO-I metrics, demonstrating superior alignment performance.
For instance, our model significantly surpasses the IP-Adapter-SDXL by 0.223 DINO-I score.
Note that IP-Adapter utilizes CLIP image embeddings for conditioning, its generated images may exhibit a bias towards CLIP's preference, potentially increasing scores when evaluated using CLIP-based metrics.
We further conduct qualitative visualization comparisons using some reference images that encompass various styles and contents. 
As presented in Figure~\ref{fig:single_ref}, our method achieves better aesthetic quality and consistency with the original image prompts.

\vspace{1mm}
\noindent\textbf{Multi-image references.}
We compare our method with IP-Adapter and the tuning-based LoRA on the MRBench in Table~\ref{tab:mrbench}. 
On the held-in split, the tuning-free EasyRef consistently achieves better performances than the tuning-based approach LoRA.
In the zero-shot setting, the results demonstrate our method surpass the IP-Adapter with embedding averaging in alignment with the reference images and user prompt.
We also present the visualizations in Figure~\ref{fig:multi_ref}.
This experiment demonstrates our framework is capable of fully mining consistent visual elements among multiple reference images while maintaining strong generalization ability.

\vspace{1mm}
\noindent\textbf{Human evaluation.}
We systematically evaluate EasyRef with IP-Adapter and LoRA in terms of reference consistency and aesthetic quality.
The human evaluation is conducted on our proposed MRBench.
Human evaluators were presented with pairwise image comparisons, one generated by EasyRef and the other by a competing model, under blind conditions to ensure fairness.
As illustrated in Figure~\ref{fig:human_eval}, EasyRef outperforms other models in both image-reference alignment and visual aesthetics in user study. 
This demonstrates EasyRef's capacity to generate high-fidelity images that conform to the provided reference images.
    
\vspace{1mm}
\noindent\textbf{Compatibility with ControlNet.}
As shown in Figure~\ref{fig:controlnet}, our EasyRef is fully compatible with the popular controllable tool, ControlNet~\cite{controlnet}.
Compared to the IP-Adapter, EasyRef can generate high-fidelity, high-quality, and more consistent results when processing multiple reference images with additional structure controls.


\subsection{Ablation Study}
\begin{figure}[tp]
    \centering
    \includegraphics[width=0.99\linewidth]{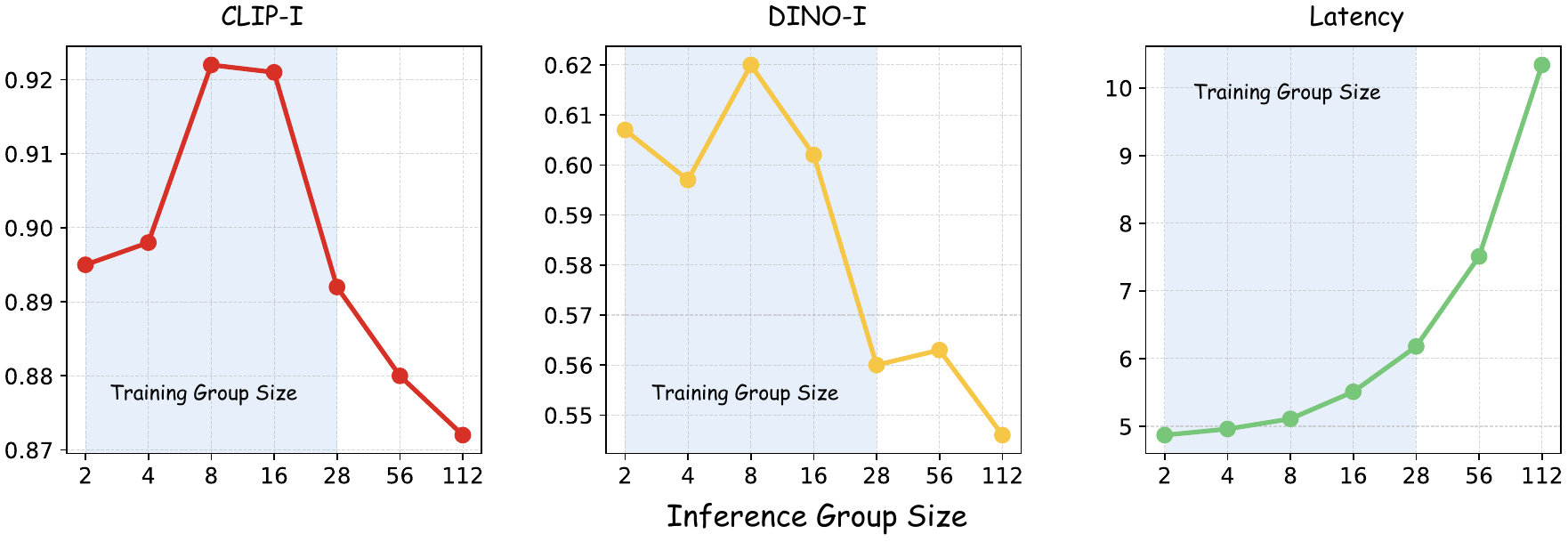}
    \caption{Evaluation of inference group size scaling. We randomly select 112 reference images and 1 target image-text pair with the same topic. Then we compute the similarities between the generated images and the target image. ``Latency'' in the figure is measured in seconds per image.}
    \label{fig:num_infer}
    \vspace{-1mm}
\end{figure}

\vspace{1mm}
\noindent\textbf{Scaling the number of reference images.}
Figure~\ref{fig:num_infer} illustrates EasyRef's performance across varying inference lengths.  
The model exhibits slightly robust performance across varying numbers of references when the number of reference images is within the training constraint.
Specifically, the performances continue to increase as the number of references increases within the training constraint.
However, performance degrades when the number of references exceeds this constraint.
This is due to the limited number of groups with more than 16 images during training and the long-context finetuning may be inadequate.
Moreover, the inference efficiency of EasyRef is further evaluated and we find it still maintains acceptable efficiency with 56 reference images due to the effecient token aggregation design.

\begin{figure}[tp]
    \centering
    \includegraphics[width=0.99\linewidth]{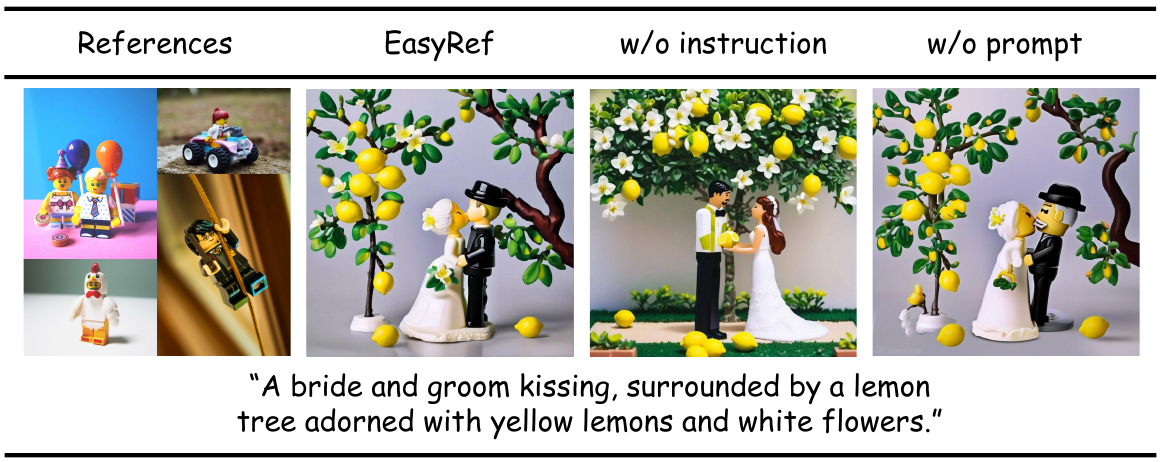}
    \caption{Impact of the multimodal instruction design.}
    \label{fig:instruct}
    \vspace{-1mm}
\end{figure}

\vspace{1mm}
\noindent\textbf{Multimodal instruction input.}
An ablation study was conducted to investigate the design of multimodal input to the LLM. 
As shown in Figure~\ref{fig:instruct}, the inclusion of instructions improves generation performance.
We speculate this leverages the MLLM's instruction-following ability to enable it to attend to crucial contents within the reference images and prompt. 
This observation is consistent with the analysis presented in LI-DiT~\cite{lidit}. 
Furthermore, incorporating the image prompt can exploit the text-understanding capacity of the MLLM and enhance text-image alignment.

\vspace{1mm}
\noindent\textbf{Reference token design.}
We first ablate the number of reference tokens on the MRBench held-out split.
The results in Table~\ref{tab:ref_token} show that too many or few tokens can hurt the performance.
Hence, we choose 64 tokens to achieve the best trade-off between accuracy and efficiency.
Furthermore, we observed comparable performances across various insertion positions (\textit{e.g.}, the final, second to last, and third to last layers of the LLM) for the reference tokens. 
Consequently, we propose to insert the reference tokens into the final layer for optimal computational efficiency.

\vspace{1mm}
\noindent\textbf{Reference aggregation design.}
In this experiment, we compare our reference token aggregation paradigm with embedding averaging and embedding concatenation.
As shown in Table~\ref{tab:ref_agg}, averaging the multi-reference representations leads to performance degradation and the concatenation can increase the reference token number by more than 5$\times$.
Therefore, utilizing the multi-image comprehension capability of MLLM can enhance the model performance.
\begin{table}[!t]
\centering
\resizebox{0.9\linewidth}{!}{
\begin{tabular}{@{}ccccc@{}}
    \toprule
    Number & Position & CLIP-I $\uparrow$ & CLIP-T $\uparrow$ & DINO-I $\uparrow$ \\  
    \midrule 
    32 tokens & -1 & 0.813 & 0.693 & 0.591 \\
    128 tokens & -1 & 0.827 & 0.705 & 0.611 \\
    \midrule
    64 tokens & -2 & 0.831 & 0.704 & 0.616 \\
    64 tokens & -3 & 0.828 & 0.702 & \textbf{0.617} \\ 
    \midrule
    \cellcolor{aliceblue}64 tokens & \cellcolor{aliceblue}-1 & \cellcolor{aliceblue}\textbf{0.833} & \cellcolor{aliceblue}\textbf{0.709} & \cellcolor{aliceblue}0.614 \\    
    \bottomrule
\end{tabular}
    }
\caption{Ablation of reference token design.}
\vspace{-2mm}
\label{tab:ref_token}
\end{table}

\begin{figure}[tp]
    \centering
    \includegraphics[width=0.99\linewidth]{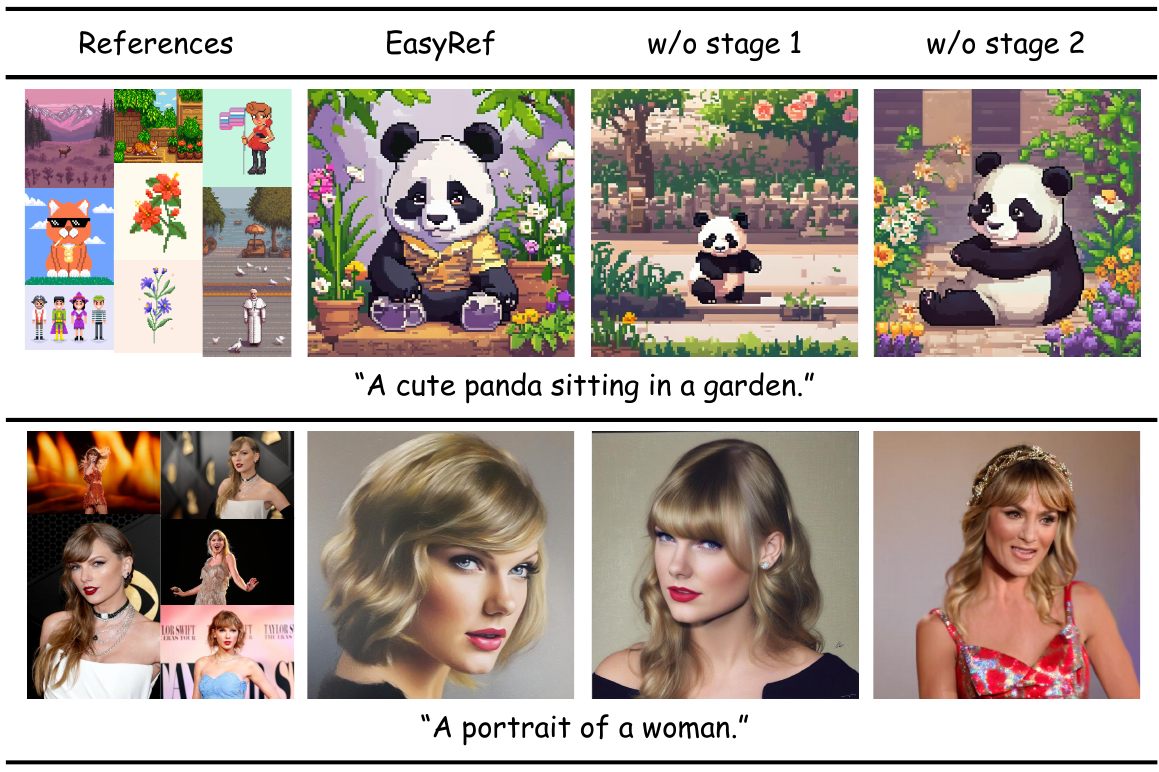}
    \caption{Effect of the progressive training scheme. ``stage 1'' and ``stage 2'' denote the alignment pretraining stage and single-reference finetuning stage, respectively.}
    \label{fig:train}
\vspace{-2mm}    
\end{figure}

\vspace{1mm}
\noindent\textbf{Progressive training scheme.}
The goal of progressive training scheme is to progressively refine the MLLM's visual capabilities, ultimately enhancing alignment performance of the diffusion model.
By systematically removing each training phase, we visualize the impact of each stage on the model's ability to capture fine-grained visual details and maintain identity consistency in Figure~\ref{fig:train}.
For some reference contents, such as the pixel art style, EasyRef without alignment pretraining or single-reference finetuning maintains comparable performance.
Only for reference images involving identity preservation (\textit{e.g.,} Taylor Swift) or complex compositions do we find significant alignment improvements when adopting all training phases. 
\begin{table}[tp]
\centering
\resizebox{1.0\linewidth}{!}{
\begin{tabular}{@{}ccccc@{}}
    \toprule
    Method & Average token number & CLIP-I $\uparrow$ & CLIP-T $\uparrow$ & DINO-I $\uparrow$ \\  
    \midrule 
    Average & 64 & 0.818 & 0.688 & 0.584 \\
    Concatenation & 354 & 0.821 & 0.692 & 0.579 \\
    \cellcolor{aliceblue}EasyRef & \cellcolor{aliceblue}64 & \cellcolor{aliceblue}\textbf{0.833} & \cellcolor{aliceblue}\textbf{0.709} & \cellcolor{aliceblue}\textbf{0.614} \\
    \bottomrule
\end{tabular}
    }
\caption{Ablation of reference representation aggregation on the MRBench held-out set. In the implementation, we average or concatenate the MLLM's representations of reference images.}
\vspace{-0mm}
\label{tab:ref_agg}
\end{table}
\section{Conclusion}
This paper presents EasyRef, a novel plug-and-play adaptation method that enables diffusion models to be conditioned on multiple reference images and the text prompt.
Our approach can effectively capture consistent visual elements within multiple reference images and the text prompt through an multi-image comprehension and instruction-following paradigm, while simultaneously maintaining strong generalization capabilities due to the integration of adapter-based injection.  
The proposed efficient reference aggregation strategy and progressive training scheme further enhance computational efficiency and fine-grained detail preservation.  
Through extensive evaluation on our newly introduced MRBench, EasyRef has demonstrably surpassed both tuning-free and tuning-based approaches, in terms of aesthetic quality and zero-shot generalization across diverse domains.  
{
    \small
    \bibliographystyle{ieeenat_fullname}
    \bibliography{main}
}

\clearpage
\renewcommand{\thesection}{\Alph{section}}
\renewcommand\thefigure{\Alph{section}\arabic{figure}}
\renewcommand\thetable{\Alph{section}\arabic{table}}
\setcounter{page}{1}
\setcounter{section}{0}
\setcounter{figure}{0}
\setcounter{table}{0}


\twocolumn[{%
\renewcommand\twocolumn[1][]{#1}%
\maketitlesupplementary
\begin{center}
    \centering
    \captionsetup{type=figure}
    \includegraphics[width=0.92\textwidth]{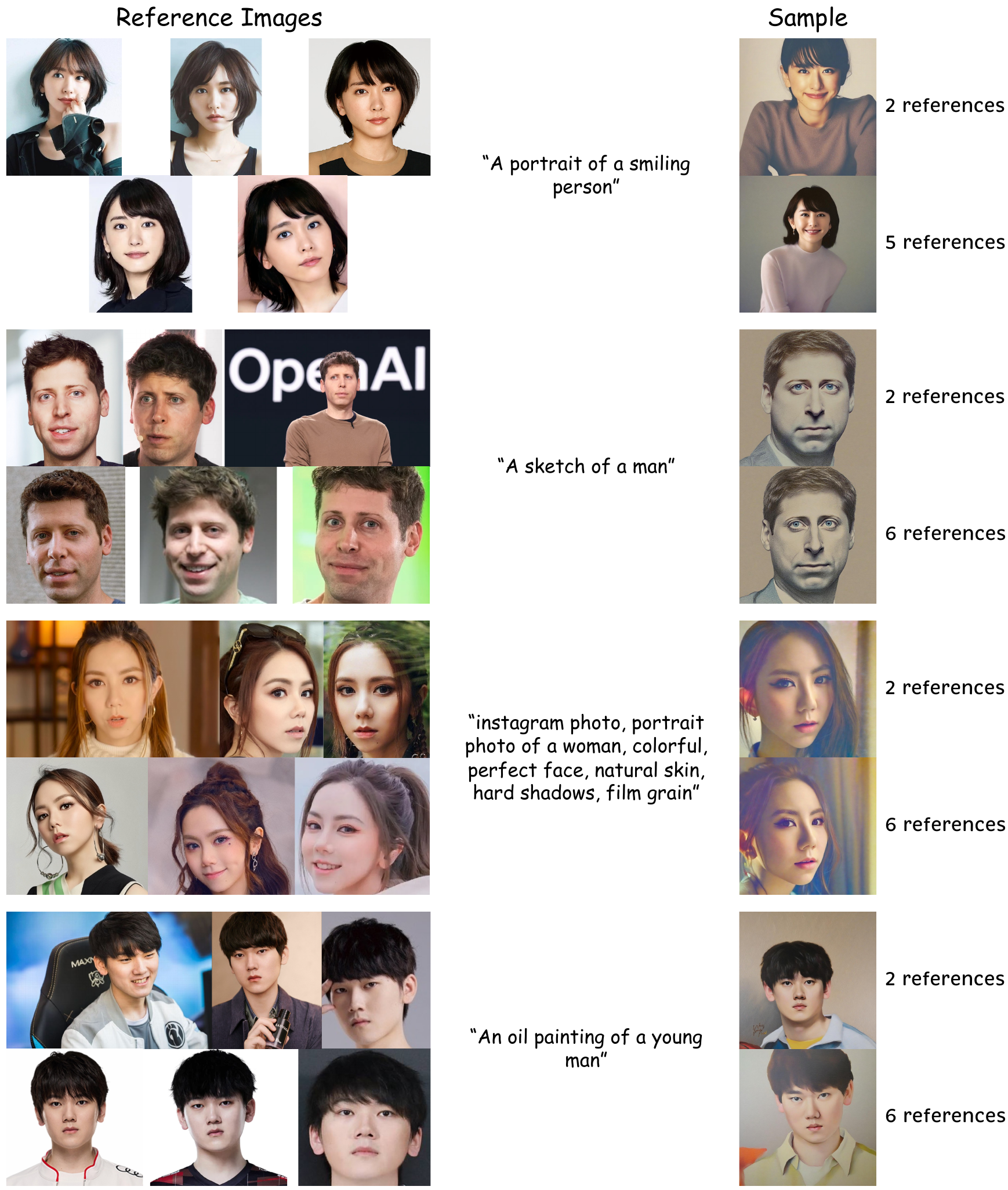}
    \caption{More generated samples of identity preservation with EasyRef in a \textbf{zero-shot setting}. We use the face images of celebrities in this experiment.}
    \label{fig:identity}
\end{center}%
}]

\begin{figure*}[tp]
    \centering
    \includegraphics[width=1.0\textwidth]{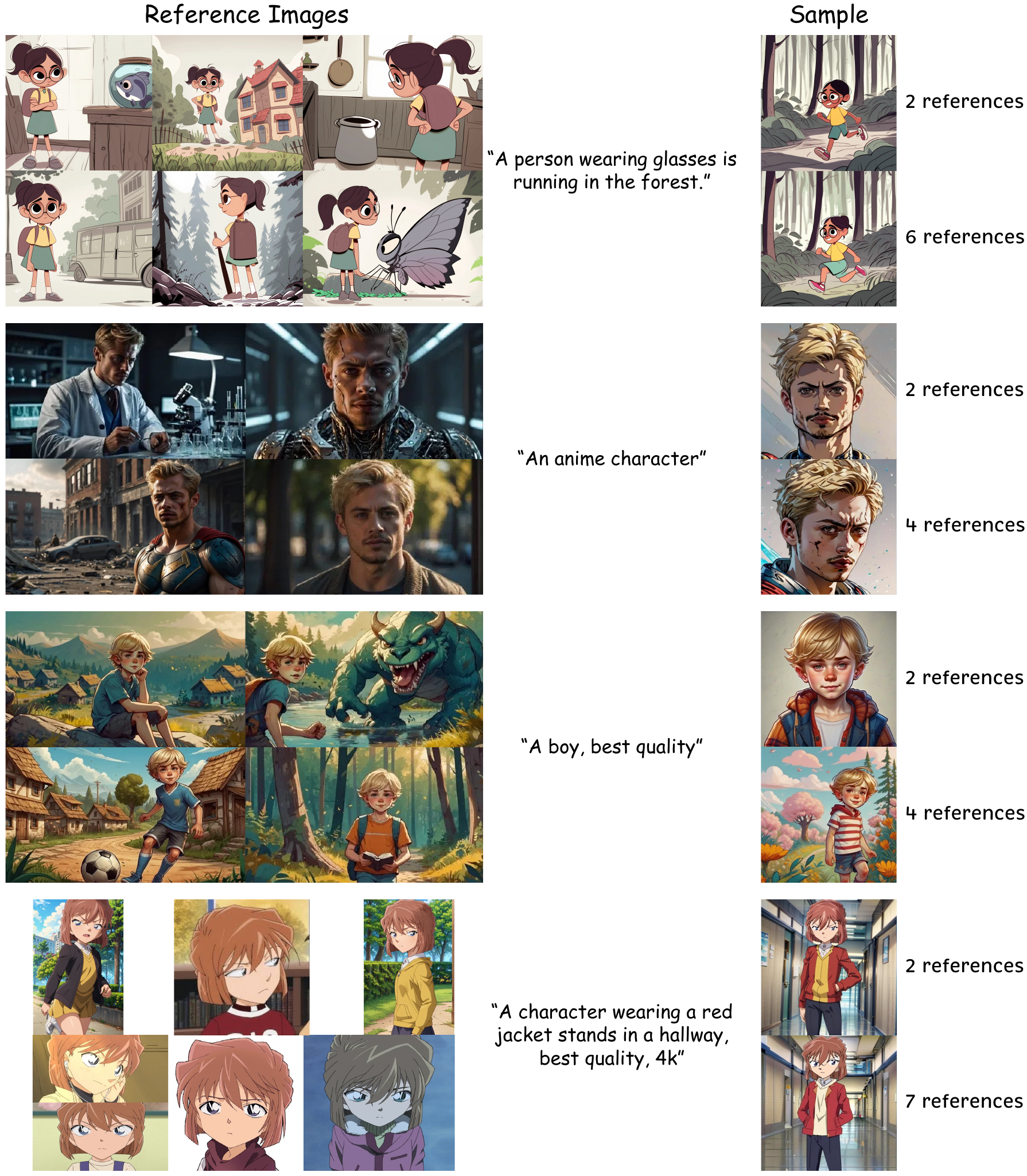}
    \caption{More generated samples of character consistency with EasyRef in a \textbf{zero-shot setting}.}
    \label{fig:character}
\end{figure*}
\begin{figure*}[tp]
    \centering
    \includegraphics[width=1.0\textwidth]{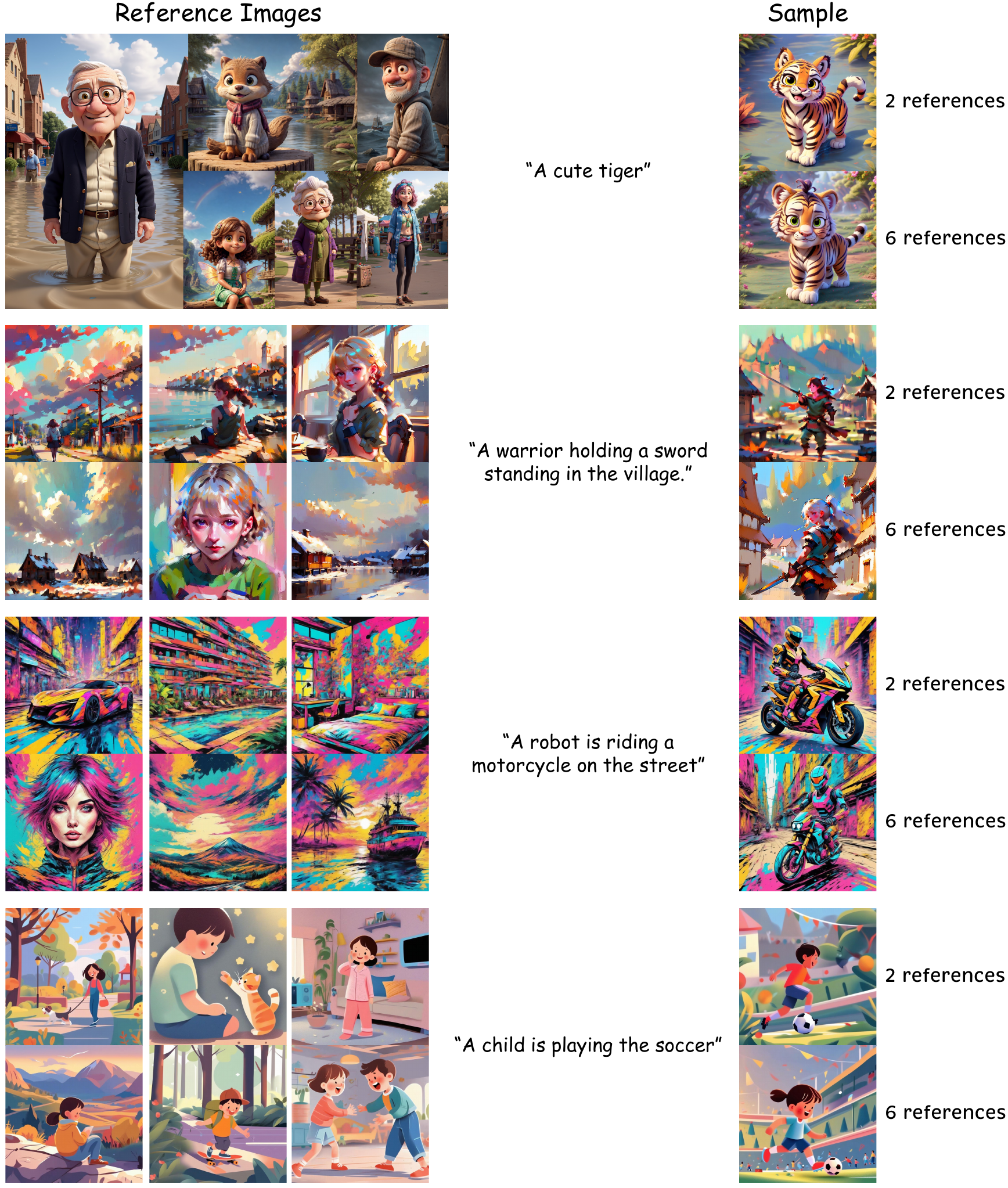}
    \caption{More generated samples of style consistency with EasyRef in a \textbf{zero-shot setting}.}
    \label{fig:style}
\end{figure*}


\end{document}